\pdfoutput=1

\documentclass[11pt]{article}

\usepackage[]{acl}
\usepackage{times}
\usepackage{latexsym}

\usepackage[T1]{fontenc}

\usepackage[utf8]{inputenc}

\usepackage{microtype}

\usepackage{inconsolata}

\usepackage{multirow}
\usepackage{booktabs}
\usepackage{graphicx}
\usepackage{pifont}
\usepackage{amsmath}
\usepackage{makecell}
\usepackage{algorithm}
\usepackage{algpseudocode}
\usepackage{subcaption}
\usepackage[normalem]{ulem}
\usepackage{comment}
\usepackage{fancyvrb}
\usepackage{arydshln}

\newcommand{\cmark}{\ding{51}}  %
\newcommand{\xmark}{\ding{55}}  %

\newcommand{\argmax}[2]{\underset{#1}{\mathrm{argmax}}\; #2}  %
\newcommand{\maxunderset}[2]{\underset{#1}{\mathrm{max}}\; #2}  %

\newcommand{\schema}[0]{\mathcal{S}}
\newcommand{\D}[0]{\mathcal{D}}  %
\newcommand{\retriever}[0]{\theta_{ret}}
\newcommand{\LM}[0]{\theta}
\newcommand{\Ek}[0]{\mathcal{E}_k}
\newcommand{\C}[0]{\mathcal{C}}  %
\newcommand{\Pool}[0]{\mathcal{P}}
\newcommand{\concat}[0]{\cdot}
\newcommand{\prompt}[0]{f_{\text{prompt}}}
\newcommand{\topP}[0]{\underset{\scalebox{.6}{\text{top-}p}}{\sim}}

\definecolor{new-pink}{rgb}{0.82, 0.02, 0.77}
\newcommand{\smalltt}[1]{\texttt{ #1}}  %
\definecolor{dst-color}{rgb}{0.2, 0.46, 0.05}
\newcommand{\dstcolor}[1]{\textcolor{dst-color}{#1}}
\definecolor{dat-color}{rgb}{0.57, 0.06, 0.77}
\newcommand{\datcolor}[1]{\textcolor{dat-color}{#1}}
\definecolor{user-color}{rgb}{0.73, 0.35, 0.01}
\newcommand{\coloruser}[1]{\textcolor{user-color}{#1}}  %
\definecolor{bstate-color}{rgb}{0.16, 0.01, 0.64}
\definecolor{prompt-comment-color}{HTML}{C60606}

\newcommand{\greencmark}{\textcolor{dst-color}{\ding{51}}} 
\newcommand{\redxmark}{\textcolor{red}{\ding{55}}}

\newcommand{\bdk}[1]{}
\newcommand{\jmf}[1]{}
\newcommand{\nilay}[1]{}
\newcommand{\cm}[1]{}
\newcommand{\gr}[1]{}

\title{Unsupervised End-to-End Task-Oriented Dialogue with LLMs:\\ The Power of the Noisy Channel}

\author{Brendan King \and Jeffrey Flanigan\\
  University of California, Santa Cruz \\
  \texttt{\{bking2,jmflanig\}@ucsc.edu}}

\begin{document}
\maketitle
\begin{abstract}
Training task-oriented dialogue systems typically requires turn-level annotations for interacting with their APIs: e.g. a dialogue state and the system actions taken at each step. 
These annotations can be costly to produce, error-prone, and require both domain and annotation expertise. 
With advances in LLMs, we hypothesize that unlabeled data and a schema definition are sufficient for building a working task-oriented dialogue system, completely unsupervised.
We consider a novel unsupervised setting of only (1) a well-defined API schema (2) a set of unlabeled dialogues between a user and agent.  We propose an innovative approach using expectation-maximization (EM) that infers turn-level annotations as latent variables using a noisy channel model to build an end-to-end dialogue agent.
Evaluating our approach on the MultiWOZ benchmark, our method more than doubles the dialogue success rate of a strong GPT-3.5 baseline.\footnote{Our code is available at \href{https://github.com/jlab-nlp/nc\_latent\_tod}{https://github.com/jlab-nlp/nc\_latent\_tod}}

\end{abstract}

\section{Introduction}

\jmf{Add: Tool use with APIs is very important, and inferring API calls from unlabeled dialogues would be very helpful}Task-oriented dialogue systems, which use APIs to complete tasks on behalf of users, have been a longstanding challenge within conversational AI.
Recent advances in large language models (LLMs) have further stimulated interest in task-oriented systems and LLMs which can use APIs as tools.
To facilitate API use, successful task-oriented dialogue systems usually employ a modular approach:
predicting a dialogue state which includes arguments to API calls, and dialogue acts for planning an appropriate response, before finally producing a natural language reply.
Training such systems typically requires expert annotation of these structured intermediates for every dialogue turn.
Even in settings where human-human dialogues are abundantly available, the high cost and expertise required to annotate the dialogues poses a significant hurdle to system development.
\begin{figure}[t]
    \centering
    \includegraphics[width=\columnwidth]{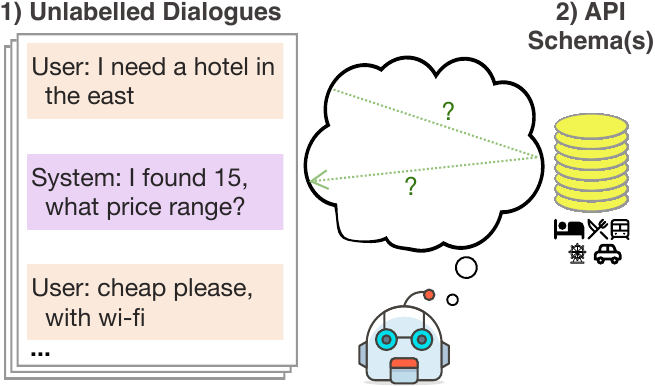}
    \caption{An overview of our unsupervised dialogue problem. We assume 1) unlabeled goal-oriented dialogues between a user and agent and 2) a well-defined schema $\schema$ with APIs suitable for fulfilling goals. We infer the unseen interactions between the agent and API, and use this to produce an end-to-end dialogue agent.}
    \label{fig:problem}
\end{figure}

Recent work has shown that LLMs can accomplish a broad set of useful tasks without any structured labels for a task \cite{brown_language_2020}. These include `zero-shot' approaches to task-oriented dialogue sub-tasks such as Dialogue State Tracking (DST) \cite{hu-etal-2022-context, king-flanigan-2023-diverse, heck-etal-2023-chatgpt}, intent detection \cite{pan_preliminary_2023}, grounded response generation \cite{li_guiding_2023}, and even zero-shot end-to-end dialogue systems \cite{hudecek-dusek-2023-large}. 
Still, existing approaches generally do not perform well enough for real-world use, and none are able to make effective use of in-domain unlabeled dialogues.

In this work, we ask whether we can use existing unlabeled dialogues (without any labels or API calls annotated) along with an API specification, to build a working dialogue agent, without needing an expert to annotate data.
This addresses a common real-world scenario. 
Many high value dialogue tasks are currently carried out by human agents, who interface a user with some software system. 
These conversations can be recorded and transcribed, and the API(s) supporting the agent typically have well-formed specifications.
However, annotating the API calls and system acts needed for aligning the two is time consuming and requires annotation expertise.
In lieu of this, `zero-shot' systems have been proposed, but these still require an expert to annotate a `formatting example' \cite{hu-etal-2022-context, king-flanigan-2023-diverse}, or a more detailed `policy skeleton' \cite{zhang-etal-2023-sgp}.

We instead propose the following setting: we assume an API schema definition $\schema$, and plenty of available human-human dialogues in natural language, but no annotations on these dialogues (\autoref{fig:problem}).
We demonstrate that one can develop a conversational agent for the API schema in this setting without any assistance from an expert annotator.  
Our contributions are as follows:
\begin{itemize}
    \item We construct an end-to-end task-oriented dialogue agent with an LLM solely from unlabeled dialogues and an API definition, without any turn-level labels or supervision from de-lexicalized utterances. To the best of our knowledge, we are the first to consider this setting.
    \item We train our dialogue system by inferring all the pseudo-labels necessary (API calls, system actions) to train a traditional end-to-end dialogue system from unlabeled dialogues, using prompts which are automatically generated from the API schema.
    \item To improve the inferred labels, we devise a novel Hard-EM \cite{dempster_maximum_1977_fixed}\jmf{fix for camera-ready} approach which uses predictions as in-context examples for the LLM and as data for iteratively fine-tuning a final model.
    \item We propose a noisy-channel `code-to-text' re-ranking approach, which is instrumental to our method and greatly improves our pseudo-label quality and final system.

\end{itemize}

\section{Preliminaries}

A task-oriented dialogue consists of turns of utterances between a user and an agent which interfaces the user with a programmable system or API to accomplish a task. Typically the system response utterance follows the user's utterance.
We denote $u_t$ as the user's utterance at turn $t$, and $r_t$ as the system's response. 
We assume the APIs supported by the system are defined in a schema $\schema$, which gives names and descriptions for all arguments supported in each API, as well as the possible values any categorical arguments may take \cite{rastogi_towards_2020}. 
This is analogous to standardized formats for API documentation, many of which could be easily converted to a schema definition.

Task-oriented systems require some method for interacting with the APIs in $\schema$. 
Modular approaches use a Dialogue State Tracking (DST) module, which predicts a belief state $b_t$\jmf{should we call this API call(s) instead?}: a collection of arguments to API call(s) needed to satisfy the user's goal. 
A belief state is commonly represented with a set of slot-value pairs:
$$
b_t = \{(s_1, v_1), (s_2, v_2), ... (s_n, v_n)\}
$$
For example, if a user says `I'm looking for a restaurant south of town', a DST system might produce the belief state \{(restaurant-area, south)\}, which can be used to query a restaurant API. We assume zero labeled belief states and infer them from unlabeled dialogues using the space of possible states supported by the schema definition $\schema$.

We also make use of system dialogue acts to structure our agent's communicative intents with a policy module. 
Given a dialogue state and context for a turn $t$, the policy predicts set of dialogue acts to be communicated in the system response $r_t$. 
For instance, the policy might determine that we should ask the user to narrow their search to a price range: $A_t = \{\text{Request(restaurant-pricerange=?)}\}$.
An appropriate system response might be: ``Sure, are you looking for a particular price range?"
Like belief states, we assume zero supervised examples of $A_t$ and infer them from unlabeled dialogues. 

\section{Method Overview}
\label{sec:methods-overview}
\begin{figure*}
    \centering
    \includegraphics[width=\textwidth]{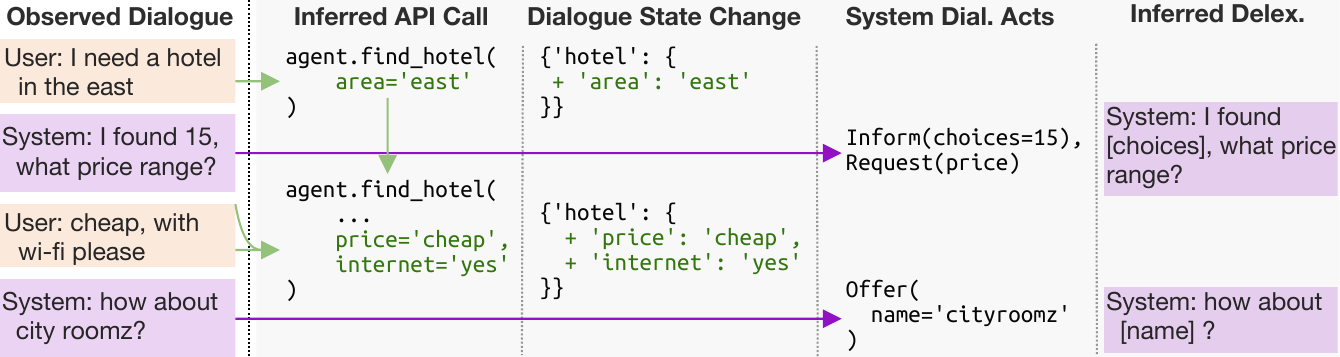}
    \caption{An overview of the latent variables annotated in our unsupervised labeling process which are used to train the dialogue model. Our \dstcolor{DST Module} (\autoref{sec:methods-dst}) infers the API call(s) with arguments at each turn, from which we can derive the dialogue state change. Our \datcolor{DAT or Act Tagging module} (\autoref{sec:methods-tagging}) predicts the dialogue acts communicated in the observed system response, which can be used to infer de-lexicalized responses for training a response generator.\jmf{We could remove the dialogue state change column, and in the text say that we convert an API call to a dialogue state for evaluation (and give an example)} \bdk{Switched positions so its API -> dialogue state. Left it in as a column to make sure people who know MWoZ can very quickly understand the API call/state mapping}}
    \label{fig:latents-overview}
\end{figure*}
We treat the turn-level labels needed for training an end-to-end dialogue system as latent variables, and infer them from unlabeled dialogues.
We assume only the fully-lexicalized sequence of user and system utterances $u_1, r_1, ... u_T, r_T$, and the schema $\schema$ defining the system's capabilities, which defines the space of valid dialogue state and act labels. 
Importantly, our prompts are automatically generated from the API schema.

In \autoref{sec:methods-offline-label}, we outline our noisy-channel prompting method for inferring the turn-level labels necessary for training our dialogue agent. We give an overview of the latent variables we infer in \autoref{fig:latents-overview}. We assume we cannot query the APIs or observe results while labeling dialogues offline, as the obtained API results may have changed.
In \autoref{sec:methods-online-system}, we train a complete dialogue agent by fine-tuning on prompts derived from our inferred pseudo-labels.

\section{Inferring Latents via Noisy Channel}
\label{sec:methods-offline-label}

In this section, we present our method for inferring latent annotations for the dialogue states $b_1...b_T$\jmf{API calls?} and dialogue acts $A_1...A_T$ for each dialogue turn $t$ given only the unlabeled user and system utterances $(u_1, r_1, u_2, r_2, ... u_T, r_T)$.
To do this, we devise a noisy-channel prompting approach for DST and dialogue act tagging (DAT) using StarCoder \cite{li2023starcoder}, a code-based LLM.
First, we use a text-to-code prompt to infer the API call(s) made by the system in each dialogue, and build the dialogue state from inferred API call arguments (\autoref{sec:methods-dst}).
We use a similar text-to-code prompt to infer the latent act(s) communicated in each agent response, so that we can reverse-engineer an agent's policy (\autoref{sec:methods-tagging}).
For both tasks, we find much better performance when re-ranking latent predictions according to a noisy-channel model, in which we condition the observed utterance on a predicted latent in a code-to-text prompt (\autoref{sec:methods-nc-prompting}).
Finally, we leverage the in-context learning ability of LLMs by re-using our predictions as exemplars (\autoref{sec:methods-retriever}).
Given these initial pseudo-labels, we iteratively improve their quality using Hard-EM \cite{dempster_maximum_1977_fixed} (\autoref{sec:methods-re-labeling}).

\subsection{Inferring API Calls and Dialogue State}
\label{sec:methods-dst}

We prompt the LLM with a text-to-code prompt for inferring the latent dialogue state as an API call.
\autoref{fig:direct-dst-prompt}  in \autoref{app:prompt-examples} gives an example of our prompt.
We generate a prompt enumerating the intents available in the schema $\schema$ as APIs callable by our agent.
Following \citet{hu-etal-2022-context}, we predict the appropriate function call conditioned on the prior system response $r_{t-1}$, the current user utterance $u_t$, and the previous belief state prediction $\hat{b}_{t-1}$. 
We then extract a dialogue state \textit{change} $\Delta \hat{b}_t$ from the arguments to the call, and compute the next dialogue state as $\hat{b}_t = \Delta \hat{b}_t + \hat{b}_{t-1}$.
While used offline here, this DST method is causal with respect to dialogue inputs and is the same as our method in online inference.

\subsection{Inferring System Acts} 
\label{sec:methods-tagging}
For inferring system acts, we use a similar text-to-code prompt for predicting the set of dialogue acts $A_t$ communicated in a given system response $r_t$.
See \autoref{fig:direct-act-tag-prompt} in \autoref{app:prompt-examples} for an example of our prompt.
We define each act our system could take in the prompt instructions.
For input from each turn, we find best performance when conditioning only on the response to tag, $r_t$.
For our set of supported acts, we use a subset of the universal dialogue acts proposed in \citet{paul19b_interspeech}, where some acts such as ``Inform'' or ``Offer'' may use slots defined in $\schema$.
For example, an agent choosing to offer to book a user at a hotel named `acorn guest house' might be represented as Offer(hotel\_name=`acorn guest house'). 
See \autoref{app:dialogue_acts} for our complete dialogue act set. 
Importantly, we use the schema definition $\schema$ and our act set to validate each act prediction, removing predicted keys which do not belong to $\schema$, or acts which are not in the set. For example, the `text' key is not valid for a `ThankYou' act, so a prediction of ``ThankYou(text=`thanks, have a good day')" would be normalized to only ``ThankYou()''.
Using the inferred system acts, we use a rule-based method to delexicalize the system responses for training the response generator (\autoref{fig:latents-overview}, right).
\jmf{maybe also expand on this in the response generation section} \bdk{added under \autoref{sec:methods-online-system}, Response Generation.}

\subsection{Noisy Channel LLM Prompting}
\label{sec:methods-nc-prompting}

We find that a noisy channel prompting method \cite{min-etal-2022-noisy} significantly improves the quality of our inferred dialogue states and acts.
Here we describe noisy channel prompting using a simple example, and then describe its application to dialogue state tracking and system act tagging.

A typical prompt for machine reading comprehension might be:
\begin{Verbatim}[fontsize=\small]
    <Optional in-context examples (c)>
    Passage: <Passage (z)>
    Question: <Question (x)>
    Answer:
\end{Verbatim}

Given this prompt of the in-context examples $c$, passage $z$, question $x$, an answer $y$ completion is found with the language model by maximizing or sampling from $Pr(y|x,z,c)$.  We call this the \textbf{direct prompt}.

The \textbf{``noisy channel'' prompt} is:
\begin{Verbatim}[fontsize=\small]
    <Optional in-context examples (c)>
    Passage: <Passage (z)>
    Answer: <Answer (y)>
    Question: <Question (x)>
\end{Verbatim}
where the likelihood of the question now depends on the answer. 
To use the noisy channel LLM prompt, we first sample $k$ samples from the direct prompt, and then pick the best output answer $y$ according to the noisy channel prompt probability. 
One can choose to score the joint probability of the answer followed by the question, i.e. $Pr(x|y,z,c)Pr(y|z,c)$, or only the conditional $Pr(x|y,z,c)$, following \citet{min-etal-2022-noisy}.\footnote{In the latter case, the prior $Pr(y|z, c)$ is uniformly $\frac{1}{k}$ for the $k$ samples from the direct prompt.}

\begin{figure}
    \centering
    \includegraphics[width=\columnwidth]{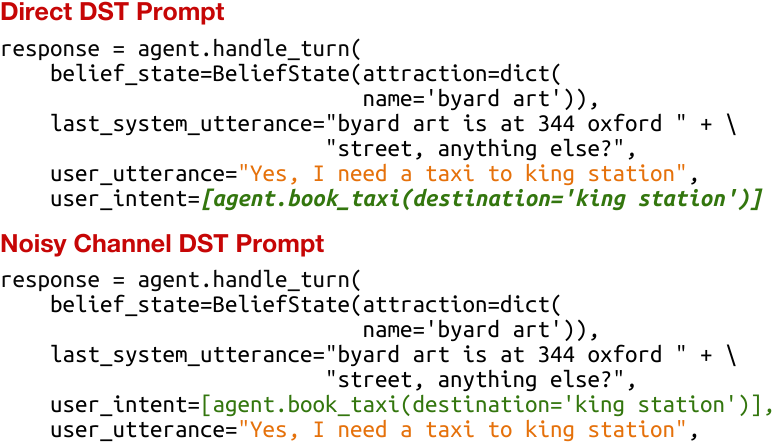}
    \caption{Instances from our `direct' and `noisy channel' prompts for DST. Best viewed in color. After sampling a \dstcolor{\textbf{\textit{DST completion}}} from the `direct' prompt, we score it by the likelihood of the input \coloruser{user utterance} conditioned on it in the `noisy channel' prompt.} 
    \label{fig:compare-direct-channel-dst}
\end{figure}

To apply this method to inferring dialogue states, we first sample a set of possible belief state changes using top-$p$ sampling \cite{Holtzman2020The} from the direct DST prompt, and then pick the best dialogue state according to the noisy channel prompt (see \autoref{fig:compare-direct-channel-dst}).  
We use an analogous procedure for inferring system acts.
For DST, we find scoring with the joint $Pr(x|y,z,c)Pr(y|z,c)$ to perform best, and scoring with the conditional $Pr(x|y,z,c)$ best for act tagging.

\subsection{Retrieval-Augmented In-context Learning}
\label{sec:methods-retriever}
To leverage the in-context learning abilities of LLMs, we retrieve from a pool of examples from our predictions.
Because we assume no labeled examples, this pool starts with zero examples and is filled incrementally.
We retrieve up to $k$ examples for in-context learning from this pool using an unsupervised dense retriever, with examples ranked by embedding cosine distance.\footnote{We use MPNet \cite{song_mpnet_2020}, available on Huggingface as \texttt{sentence-transformers/all-mpnet-base-v2}} 
We use $k=8$ and $k=6$ for DST, DAT respectively. 
For retriever inputs, we use $(\hat{b}_{t-1} \concat r_{t-1} \concat u_t)$ and $(u_t \concat r_t)$ for DST and DAT respectively, where $\concat$ indicates concatenation.
Applied naively, this in-context learning approach can suffer a majority label bias \cite{pmlr-v139-zhao21c}. We adjust for biases introduced in the initially small example pool by 1) not using any in-context examples until we have a minimum of $n=32$ examples in the pool and 2) using our API schema $\schema$ to require at least $4$ distinct labels in each set of in-context examples.\footnote{We consider two dialogue state change labels to be distinct if they update different \textit{slots}, and two act labels to be distinct if they embody different acts or different slots}
Our algorithm for producing initial pseudo-labels is in \autoref{app:offline_algorithm}.
\subsection{Refining the Labels with Hard-EM}
\label{sec:methods-re-labeling}

While the labels we produce in \autoref{sec:methods-dst}-\autoref{sec:methods-retriever} can be used directly for training an end-to-end dialogue system, we find their quality can be improved through expectation-maximization \cite{dempster_maximum_1977_fixed}.
For every dialogue turn in our dataset, our initial pseudo-labels provide the expected dialogue state and system dialogue acts according to our zero-shot system.
We then jointly fine-tune an LLM as a noisy-channel DST \& DAT system to maximize the likelihood of these expected labels.
We use smaller version of our prompted LLM, StarCoder 3B \cite{li2023starcoder}.

For each turn, we derive (prompt, completion) pairs for `direct' text-to-code and `channel' code-to-text DST and DAT modules, as defined in \autoref{sec:methods-offline-label}.
We then combine and shuffle these pairs into a single training set for joint fine-tuning.
For efficient training, we shorten our prompts by removing in-context examples as well as the function definitions used in the in-context learning setting. 
We find up-sampling the `channel' prompts so that there is a 2:1 ratio of `channel' to `direct' instances for training improves performance. 

After fine-tuning, the model can be used to produce improved pseudo-labels by re-labeling each dialogue, using the same noisy-channel inference methods. Following this, we can repeat the fine-tuning process.
This train and re-label process can be repeated for any number of iterations, though we find a single re-labeling is sufficient.

\section{End-to-End System}
\label{sec:methods-online-system}

Following \cite{su-etal-2022-multi}, we utilize a multi-task fine-tuning method for training a single LLM as a complete dialogue system, consisting of a dialogue state tracker, policy, and response generator.

\paragraph{DST} For the DST sub-task, we again use both `direct' and `channel' (prompt, completion) pairs. 
This allows us to use the same noisy-channel inference method presented in \autoref{sec:methods-offline-label}.

\paragraph{Policy} For the Policy sub-task, we use a text-to-code prompt where we simply condition on the $k$=5 most recent utterances in the dialogue history: $H_t = (u_{t-2}, r_{t-2}, u_{t-1}, r_{t-1}, u_{t})$. 
The completion is the current turn's system acts $A_t$, which will be used to ground the next response $r_t$. 
We do not use a noisy-channel variant for Policy, and greedily decode an act prediction at inference time:

$$
\hat{A}_t = \argmax{A_t \in \mathcal{V}^*}{P(\prompt(H_t)))}
$$

\paragraph{Response Generation} For Response Generation, we condition on the turn's observed system and user utterances $(r_{t-1}, u_t)$ and our policy's act prediction $\hat{A}_{t})$. The completion is the observed system response $r_t$. We also do not use a noisy-channel variant for response generation, and greedily decode the response:
$$
\hat{r}_t = \argmax{A_t \in \mathcal{V}^*}{P(\prompt(r_{t-1}, u_t, A_t)))}
$$
Following prior works, we predict \textit{delexicalized} responses, where values for slots in the system response are replaced with placeholders for the slot name. 
For example, instead of generating ``The phone number for acorn guest house is 555-5309'' directly, we would predict ``The phone number for the [hotel\_name] is [hotel\_phone]'', where values could be filled in.
Importantly, we never presume access to gold delexicalized responses.
Instead, we use our predicted acts, e.g. "Inform(name=`acorn guest house', phone=`555-8309')" to delexicalize the observed response for training.

\paragraph{End-to-end Training}
Our approach to end-to-end training is as follows. 
We first derive (prompt, completion) pairs for each module (DST, Policy and Response Generation), including `channel' prompts for DST.
To improve training efficiency, we shorten each prompt by removing in-context examples and function definitions from the in-context learning setting.
We then combine and shuffle these pairs into a single training set for joint fine-tuning and up-sample the `channel' prompts for DST using the same 2:1 ratio. 
Using this training set, we fine-tune StarCoder 3B using cross-entropy loss and AdamW with default hyperparameters. 

\begin{table*}[ht!]
\centering
\resizebox{\textwidth}{!}{
\begin{tabular}{lrrr|rrrr}
\textbf{Model} & \textbf{Schema?} & \textbf{Labels?} & \textbf{Dialogues?} & \textbf{Inform} & \textbf{Success} & \textbf{BLEU} & \textbf{Combined} \\
\hline
\multicolumn{8}{c}{\textbf{Supervised Results}} \\
\hline
PPTOD \cite{su-etal-2022-multi} & \cmark & \cmark & \cmark & 82.6 & 72.2 & 18.2 & 95.6 \\
DiactTOD \cite{wu-etal-2023-diacttod} & \cmark & \cmark & \cmark & 89.5 & 84.2 & 17.5 & 104.4 \\
Our (supervised) & \cmark & \cmark & \cmark & 67.9 & 61.7 & 14.6 & 79.4 \\ 

\hline
\multicolumn{8}{c}{\textbf{Zero-Shot with Formatting Example(s)}} \\
\hline
SGP-TOD-GPT3.5 \cite{zhang-etal-2023-sgp} & \cmark & Few (\ddag) & \xmark & 82.0 & 72.5 & 9.22 & 86.5 \\
\hline
\multicolumn{8}{c}{\textbf{Fully Unsupervised Results}} \\
\hline
\multicolumn{8}{l}{\textit{\textbf{Sees gold delexicalized conversation history}}} \\
LLaMa\textsuperscript{\textdagger} & \cmark & \xmark & \xmark & - & 4 & 1.61 & - \\
GPT 3.5 Turbo\textsuperscript{\dag} & \cmark & \xmark & \xmark & 44.8 & 31.2 & 3.3 & 41.3 \\
\hdashline
\multicolumn{8}{l}{\textit{\textbf{Sees only fully-lexicalized dialogues}}} \\
GPT 3.5 Turbo ($-$ gold delex.) & \cmark & \xmark & \xmark & 40.7 & 26.7 & 3.7 & 37.4 \\
Ours (StarCoder 15B - no EM) & \cmark & \xmark & \xmark & 50.0 & 19.6 & 3.2 & 38 \\
Ours (StarCoder 3B - w/ EM) & \cmark & \xmark & \cmark & \textbf{78.1} & \textbf{68.3} & \textbf{13.6} & \textbf{86.8} \\
\hline
\end{tabular}
}
\caption{Unsupervised end-to-end results in MultiWOZ 2.2. (\dag) indicates models from \citet{hudecek-dusek-2023-large}. Results for LLaMa are from \citet{hudecek-dusek-2023-large}, which does not report the Inform rate. 
(\ddag) SGP-TOD uses a prompt with both a formatting example and a ``Policy Skeleton'', which contains an additional 10-20 hand-crafted instances of the correct system acts and response for an input user utterance or returned DB result. For fairer comparison in our fully unsupervised setting, we re-run the GPT 3.5 baseline without the supervision of de-lexicalized responses provided in the conversation history ($-$ gold delex.). Despite far fewer parameters, we find substantial improvements in our methods which leverage unlabeled dialogues}
\label{tab:main-results}
\end{table*}

\section{Experiments}

We conduct unsupervised end-to-end dialogue (E2E) and dialogue state tracking (DST) experiments on the MultiWOZ 2.2 dataset \cite{zang_multiwoz_2020, budzianowski2018large}, containing over ten thousand multi-domain task-oriented dialogues crowd-sourced in a wizard-of-oz setup. We use the fully lexicalized, unlabeled dialogues from the training set to build our system, and evaluate on the test set.
First, we demonstrate the value of our approach in an end-to-end dialogue evaluation, following prior works on task-oriented dialogue (\autoref{sec:expts-e2e}). 
Second, we conduct a dialogue state tracking evaluation to more carefully evaluate the quality of our pseudo-annotations (\autoref{sec:expts-dst}).

\subsection{End-to-End (E2E) Experiments}
\label{sec:expts-e2e}

In E2E experiments, we use our complete system to both predict API call arguments and generate a next system response in natural language. 
We evaluate our generated responses with Inform rate, Success rate, and BLEU, as well as a Combined score of $0.5(\text{Inform} + \text{Success}) + BLEU$, following prior works \cite{budzianowski2018large, nekvinda-dusek-2021-shades}. We provide details on these metrics in \autoref{app:metric_details}.

We compare our approach to the previous state-of-the-art unsupervised methods, a GPT-3.5 zero-shot baseline \cite{hudecek-dusek-2023-large}, and SGP-TOD \cite{zhang-etal-2023-sgp}. 
Where possible, we report results for both the original approach and modifications required to fit our fully unsupervised setting.
For reference, we also run our own method in the fully-supervised setting. 
We train a model using the procedure in $\autoref{sec:methods-online-system}$ using the annotations sourced from crowd-workers in the MultiWOZ 2.2 corpus \cite{budzianowski2018large, zang_multiwoz_2020}, rather than the pseudo-labels predicted in \autoref{sec:methods-offline-label}.
We also compare with existing supervised approaches as a reference point. 
We include DiactTOD \cite{wu-etal-2023-diacttod}, which to our knowledge is the supervised state-of-the-art, and PPTOD \cite{su-etal-2022-multi}, which uses a multi-task fine-tuning approach similar to our own in \autoref{sec:methods-online-system}, for T5 encoder-decoder models \cite{raffel_exploring_2020}.

\subsection{DST Experiments}
\label{sec:expts-dst}

We conduct multi-domain DST experiments on the MultiWOZ Dataset in order to evaluate the quality of our pseudo-annotations. We use our DST Module to predict and evaluate only latent dialogue states, which collect the arguments required for unseen API calls.

Following prior works, we evaluate DST performance with joint-goal accuracy (JGA), or whether a given dialogue state is completely accurate. More details are available in \autoref{app:metric_details}.

We compare to our ChatGPT 3.5 Turbo baseline \cite{hudecek-dusek-2023-large}, as well as prior zero-shot DST methods.
These include IC-DST \cite{hu-etal-2022-context}, which re-frames DST as text-to-SQL, and RefPyDST which re-frames DST as text-to-python \cite{king-flanigan-2023-diverse}. By default, both of these works use OpenAI Codex \cite{chen_evaluating_2021} which is now unavailable. We apply their prompting approaches to StarCoder 15B for a clearer comparison.

\begin{table}[]
    \centering
    \resizebox{0.95\columnwidth}{!}{
    \begin{tabular}{l|r}
        \hline
         \multicolumn{2}{c}{\textbf{With One Formatting Example}} \\
         \hline
         IC-DST (StarCoder 15B) & 24.58 \\
         RefPyDST (StarCoder 15B) & 17.17 \\
         IC-DST (Codex) & 35.02 \\
         RefPyDST (Codex) & 40.88 \\
         \hline
         \multicolumn{2}{c}{\textbf{Fully Unsupervised}} \\
         \hline
         IC-DST (StarCoder 15B) & 15.66 \\
         RefPyDST (StarCoder 15B) & 13.88 \\
         GPT 3.5 Turbo \cite{hudecek-dusek-2023-large} &  13.05 \\
         Ours (StarCoder 15B $\rightarrow$ 3B) & \textbf{39.70} \\ 
    \end{tabular}
    }
    \caption{Joint Goal Accuracy (JGA) of our method's dialogue state predictions and zero-shot baselines. 15B $\rightarrow$ 3B indicates our approach uses StarCoder 15B to compute initial labels (\autoref{sec:methods-dst} - \autoref{sec:methods-retriever}) and StarCoder 3B when iteratively fine-tuning and re-labelling (\autoref{sec:methods-re-labeling}), for a final model size of 3B}
    \label{tab:zero-shot-multi}
\end{table}

\section{Results}
\label{sec:results}
\subsection{E2E Performance} 
We present E2E results for our unsupervised dialogue agent in \autoref{tab:main-results}. 
We find that our method achieves state-of-the-art performance in our fully unsupervised setting, more than doubling the Success Rate and Combined score of the GPT 3.5 Turbo baseline of \citet{hudecek-dusek-2023-large}.
When we remove the supervision of delexicalization for fairer comparison ($-$ gold delex.), we find even greater improvement across all end-to-end metrics.
As discussed in \autoref{sec:related-work}, SGP-TOD uses both a supervised formatting example and a `Policy Skeleton', containing additional supervision for Policy and Response Generation.
With no implementation publicly available, we were unable to run a modified version of their experiments without this supervision for fair comparison. 
Despite a less-supervised setting, our method is able to perform comparably, even slightly out-performing SGP-TOD in Combined score.
Remarkably, our unsupervised EM approach also outperforms the supervised variant of our model due to improvements in Inform and Success rate, suggesting the Dialogue acts we infer are of high quality. 
\subsection{DST Performance} 
Our DST results are shown in \autoref{tab:zero-shot-multi}. Where possible, we distinguish between `zero-shot' results which include a hand-engineered formatting example, and the same method applied without the formatting example.\footnote{Due to the deprecation of OpenAI Codex, we were unable to run experiments for IC-DST or RefPyDST without a formatting example on the original Codex model}
We find that our method significantly outperforms our GPT 3.5 Turbo baseline by 26\% joint goal accuracy. 
Our approach performs nearly as well as the best method using OpenAI Codex with a supervised formatting example, using less than 10\% of the parameters at any time (175B vs. 15B). 
When applying the IC-DST and RefPyDST prompting methods to StarCoder, our method significantly outperforms both, with and without a formatting example.

\subsection{Ablations} 
\begin{figure}
    \centering
    \includegraphics[width=\columnwidth]{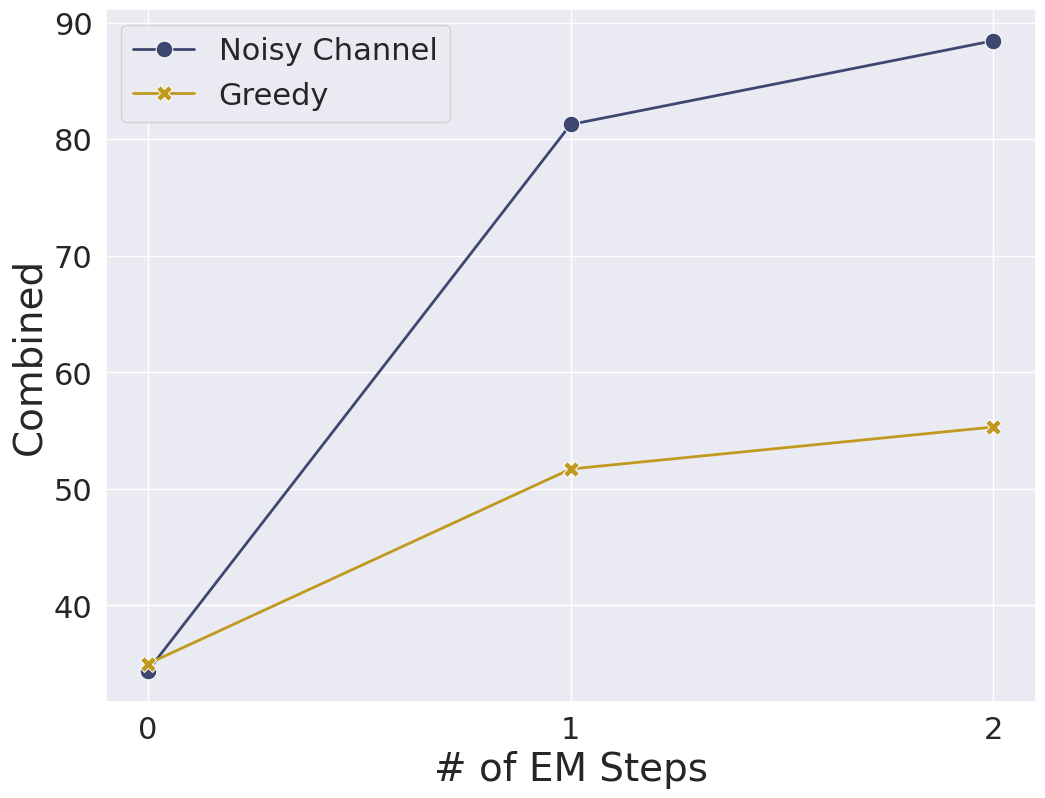}
    \caption{Combined score ($0.5(\text{Inform} + \text{Success}) + BLEU$) vs. the number of steps of expectation-maximization in our Noisy Channel method vs. a Greedy Ablation. `0' is zero-shot inference}
    \label{fig:success-v-steps}
\end{figure}

In \autoref{fig:success-v-steps}, we conduct an ablation to evaluate both the impact of our noisy channel modeling and the value of iterative re-labeling in our EM approach.
We compare our proposed system to one in which each module is replaced by only greedily sampling from its `direct' variant, at both labeling and end-to-end inference time. We plot our Combined end-to-end performance across iterations of EM, with `0' indicating our zero-shot system.
We find that EM improves our end-to-end performance in both our noisy-channel approach and greedy ablation, and that our noisy-channel inference methods are important to dialogue success, with a 30 and 33 point improvement over our greedy baseline with 1 and 2 EM steps, respectively. 
We report Inform, Success, BLEU, and joint goal accuracy (JGA) for this ablation in \autoref{app:additional-ablations}.

\begin{table}[]
    \centering
    \resizebox{\columnwidth}{!}{
    \begin{tabular}{l|rr}
         \textbf{Error Type} & \textbf{Noisy Channel} & \textbf{Greedy} \\
         \hline
         Successful (no errors) & 5 & 0 \\
         \hdashline
         DST Failure: Incorrect State & 1 & 1\\ 
         Policy Failure: Inappropriate Dialogue Act & 0 & 8 \\ 
         Policy Failure: Suboptimal Dialogue Act & 4 & 2 \\ 
         Response not faithful to act & 1 & 6 \\
    \end{tabular}
    }
    \caption{An analysis of errors occurring in a human evaluation of 10 validation dialogues, comparing our noisy channel model to a greedy ablation. Any one error can make the dialogue unsuccessful, but more than one error can occur. A suboptimal dialogue act is acceptable in the context and differs substantially from the gold human dialogue act.}
    \label{tab:error-analysis-summary}
\end{table}

\subsection{Error Analysis}
\label{sec:error-analysis}
\begin{figure}
    \centering
    \includegraphics[width=\columnwidth]{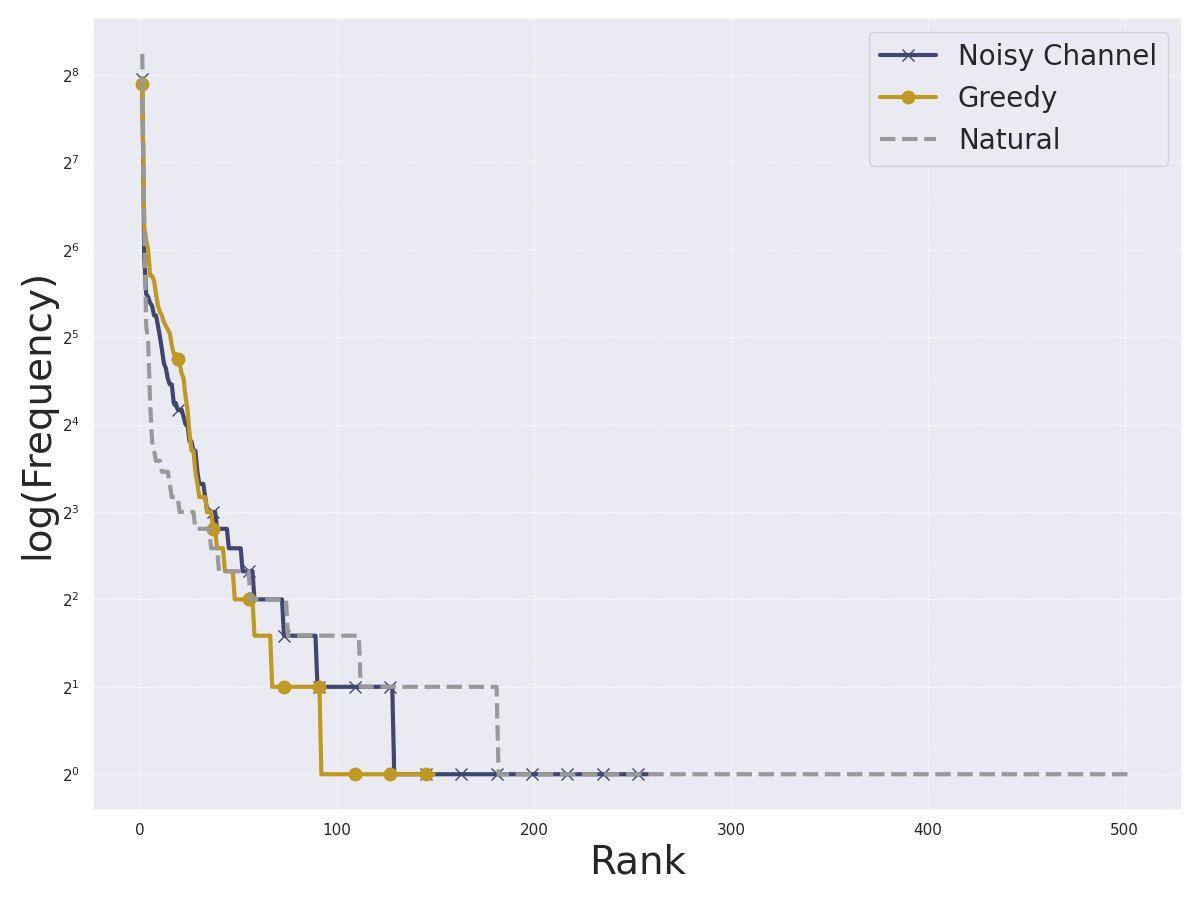}
    \caption{log(Frequency) vs. Rank of dialogue acts used by each model over a 200 dialogue sample of the validation set. `Natural' refers to human annotations. We find our Noisy Channel approach uses a higher number of unique dialogue acts than the Greedy approach and better matches the characteristics of the distribution used by human annotators}
    \label{fig:rank-freq}
\end{figure}

We conduct an error analysis comparing the greedy model to our best noisy channel model (both with two EM steps). 
We randomly sampled 10 validation dialogues where at least one model was unsuccessful and analyze each system's behavior.  Results of our human evaluation are shown in \autoref{tab:error-analysis-summary}, and examples of each error type can be found in \autoref{app:error-analysis}.  A common mistake (8 out of the 10 dialogues) of the greedy model is predicting incorrect dialogue acts and then producing a response that is not grounded in the dialogue. 
We find the noisy channel model does not make this type of mistake in any of the 10 dialogues.
We hypothesize that the greedy model overuses frequently occurring dialogue acts like `Request' regardless of context, and evaluate this by plotting the rank vs. frequency of dialogue acts not considering the values\footnote{For the rank vs. frequency analysis, we consider the dialogue act for a turn to be the set of acts and slots (but not values) used by the system. For example the act ``Inform(hotel\_name=`acorn guest house')" becomes \{(Inform, hotel\_name)\}} for each model in \autoref{fig:rank-freq}. 
We find that while both policies overuse common dialogue acts relative to a human agent, our noisy-channel method better utilizes the long tail of possible dialogue acts.

\section{Contamination Analysis}
\label{sec:contamination-analysis}

Evaluation of unsupervised methods that use LLMs has the potential issue of \textbf{task contamination}, where supervised examples are seen in pretraining data \cite{li_task_2024}.  Inclusion of supervised examples of the task in LLM pretraining data would render the model no longer unsupervised and the evaluation potentially biased: tasks for which the training data has been seen may have a higher performance than truly unsupervised tasks.

To address this issue, we quantify the presence of contamination in LLM pre-training data and estimate its potential impact on our results.
Fortunately, the complete pre-training corpus for StarCoder is publicly available for analysis.\footnote{\href{https://huggingface.co/datasets/bigcode/starcoderdata}{https://huggingface.co/datasets/bigcode/starcoderdata}}

\begin{table}[]
    \centering
    \resizebox{\columnwidth}{!}{
    \begin{tabular}{l|r:rr}
         \textbf{Task} & \textbf{Turns} & \textbf{Correct} & \textbf{Authentic} \\
         \hline
         Act Tagging & 42  & 21 & 5 \\  
         DST & 42 & 36 & 19 \\  
    \end{tabular}
    }
    \caption{Number of discovered contaminated turns per task, as well as the number which are correct or verified as being in the MultiWOZ dataset. The MultiWOZ dataset contains $71,522$ labeled turns for each task, indicating fewer than $0.06\%$ are found to be contaminated}
    \label{tab:contamination}
    
\end{table}

We conduct an exhaustive search for supervised pairs of our dialogue subtasks in the StarCoder pretraining data using a semi-automated search with manual review.  Details of our search procedure are in \autoref{app:contamination_details}.  We find no complete dialogues with supervised labels.  We do find 42 turns labeled with act tagging, and 42 turns labeled with DST in the pre-training corpus, categorized in \autoref{tab:contamination}.\footnote{The average dialogue length in MultiWOZ is 13.9 turns. Put together, the set of contaminated turns would be roughly the length of 6 dialogues}
We consider a $(x, y)$ pair to be `Correct' if the state change/dialogue act $y$ is actually correct for the utterance $x$, and to be `Authentic' if the $(x,y)$ pair is found verbatim in the MultiWOZ corpus.\footnote{A `Correct' pair might arise from printing training data, and an incorrect pair from discussion of a failure case.}
Astonishingly, we find half of the found Act Tagging pairs are incorrect, and could possibly mislead a pre-trained model if the model learned from them.
We also find that less than half of the turns are authentic for either task, and find a number of them derive from Github issues discussing problems with dialogue simulators.

Additionally, we estimate the degree to which the contamination we discover could exaggerate expected performance of our method on an unseen schema, by using contaminated $(x, y)$ pairs as in-context examples.\footnote{Ideally, one would pre-train an identical StarCoder model on a corpus \textit{without} contamination. However, this is computationally impractical. 
We are also not aware of any available LLM that can be verified as not contaminated for this task.}

In \autoref{tab:contaminated_experiment}, we compare our zero-shot prompt, which receives no examples of any kind, with a `contaminated' variant which uses $k$=3 examples derived from contamination in the pre-training corpus.
The `contaminated' model retrieves the most relevant contaminated fragments from a pool using the dense retrieval approach described in \autoref{sec:methods-retriever}. 
These are inserted as a triple-quoted string block, so that the prompt remains syntactically valid python.
By leaving contaminated examples in their original format, we test whether their inclusion elicits memorized knowledge rather than providing guidance on input/output formatting.
Surprisingly, we find including this supervision via contaminated fragments \textit{hurts} performance, indicating that these examples do not provide meaningful supervision for our task.
Further, the substantial gains in our noisy-channel EM approach suggest our method is doing more than simply eliciting schema-specific knowledge memorized in pre-training.
 
\begin{table}[]
    \centering
     \resizebox{\columnwidth}{!}{
    \begin{tabular}{l|rrrr}
    \textbf{Method} & \textbf{Inform} & \textbf{Success} & \textbf{BLEU} & \textbf{Combined} \\
       \hline
       Ours (zero-shot) & \textbf{49.0} & \textbf{15.0} & 3.0 & \textbf{35.0} \\
       Ours (k=3 contam ex.) & 44.5 & 14.0 & \textbf{3.8} & 33.1 \\
       \hline
       Ours (Full EM) & \textbf{80.5} & \textbf{69.0} & \textbf{13.7} & \textbf{88.5} \\
    \end{tabular}
    }
    \caption{Performance comparison when we include contaminated in-context examples. We find \textit{including} this supervision hurts performance, and does not explain the strong performance of our noisy-channel EM approach}
    \label{tab:contaminated_experiment}
\end{table}

\section{Related Work}
\label{sec:related-work}

\paragraph{Zero-shot Dialogue} A few recent works have proposed zero-shot approaches to dialogue problems using LLMs. 
\citet{hu-etal-2022-context} and \cite{king-flanigan-2023-diverse} propose DST methods which prompt code based LLMs in a text-to-SQL or text-to-program format, respectively.
Similarly, \cite{li-etal-2024-large-language-models} propose a prompt-based function-calling approach to DST using LLMs.
Each of these methods rely on prompts tailored to the schema and the use of one or more supervised `formatting' example(s), which requires annotation expertise. 
\citet{zhang-etal-2023-sgp} extends this approach to end-to-end task-oriented dialogue by adding a policy prompter for GPT 3.5. 
In addition to a formatting example, their policy prompt requires a hand-crafted `policy-skeleton' consisting of examples of the appropriate system act and reply in response to different user utterances or database results.
Our approach differs in that we require zero labeled examples of any kind. 
\citet{hudecek-dusek-2023-large} propose a zero-shot end-to-end method for prompting instruction-tuned LLMs like GPT 3.5.
However, this method presumes \textit{delexicalized} system responses $r_1 ... r_{t-1}$ in the conversation history as input, where entities are replaced with placeholders. 
Producing these inputs requires ground-truth annotations and gives a form of supervision about the entities and their attributes within a dialogue (see \autoref{tab:main-results} for a comparison for  GPT 3.5 Turbo with and without delex supervision).
In contrast, we only assume fully-lexicalized dialogues, which do not provide this supervision and require no human annotation.
We adapt the method of \citet{hudecek-dusek-2023-large} to use lexicalized dialogues as inputs, and use this approach as our baseline.
\citet{chung-etal-2023-instructtods} propose an end-to-end method which prompts GPT-4 for interactions with a knowledge base before producing a response, however it generalizes poorly to the multi-domain setting.

\paragraph{Semi-supervised TOD} Some works propose semi-supervised approaches to end-to-end task-oriented dialogue. 
\citet{zhang_probabilistic_2020} propose an end-to-end sequence-to-sequence model where the dialogue state is a latent variable. 
\citet{liu_variational_2021} adapt this approach for use with pre-trained language models, fine-tuning GPT-2.
While successful, these approaches require a non-trivial amount of supervised data.
\citet{hudecek-dusek-2022-learning} learn latent dialogue acts using variational recurrent neural networks without turn level dialogue state labels, but still require observed DB/API calls and responses.
Other semi-supervised works also evaluate their method in an unsupervised setting \cite{jin_explicit_2018, liu_unsupervised_2023}. 
However, these works also assume delexicalized training dialogues, which requires ground-truth annotation and gives a form a supervision to the model.

\paragraph{Noisy channel and re-ranking methods} 
A few previous works have utilized noisy channel methods for task-oriented dialogue or prompting methods. \citet{liu_pretraining_2021} pre-train a noisy channel for task-oriented dialogues as a sequence to sequence model, requiring significant labeled training data. \citet{min-etal-2022-noisy} propose noisy channel prompting for few-shot classification, which inspires our generalization to the generative setting.

\section{Conclusion}

We introduce the first (to our knowledge) approach for building a working task-oriented dialogue system with large language models using only unlabled dialogues and an API schema.
The approach leverages an LLM and expectation-maximization to infer missing labels as latent variables from unlabeled dialogues. 
In ablations, we find that a noisy channel approach vastly improves performance, and explore some possible explanations for this in an error analysis.  This work advances the state-of-the-art for using LLMs to build task oriented dialogue systems with minimal annotations.  Future work can consider applying our method for inferring structured representations from natural text to other NLP tasks.

\section{Limitations}

Data contamination in LLM pre-training poses a hurdle for accurate benchmarking across NLP, and particularly for unsupervised methods.
In an idealized setting, there would be a suitably strong task-oriented dialogue benchmark that could be verified as not belonging to the pre-training corpus of each new and more capable LLM.
This is not the case for our setting or for many others, and warrants careful attention from the NLP community.
For our setting, we were able to properly define problematic contamination and search for it in our LLM's pre-training corpus, thanks to the open release of the pre-training data.
We found limited contamination and demonstrated that the contamination we found was not helpful in eliciting task knowledge that might have been memorized in pre-training. 

All experiments in this paper were conducted on pre-existing public dialogue corpora, collected explicitly for training task-oriented dialogue agents with the knowledge of all participants \cite{budzianowski2018large}. 
Our use of the StarCoder model also falls within the terms of it's Responsible AI License.
It is important that subsequent applications of our method also adhere to any fair-use policies governing collected dialogues or transcripts.

\bdk{Points to discuss: fair access and use of unlabeled dialogues}

\section*{Acknowledgements}
We thank Nilay Patel, Geetanjali Rakshit, Zekun Zhao, Brian Mak, Changmao Li, Rongwen Zhao, and other JLab members for insightful feedback on preliminary drafts and discussions of this work. 
We also thank the anonymous reviewers and area chairs for their detailed and helpful feedback. 
The authors were supported in part by the NSF National
AI Institute for Student-AI Teaming (iSAT) under
grant DRL 2019805. The opinions expressed are
those of the authors and do not represent views
of the NSF. 
We are thankful for the computing resources provided by the Pacific Research Platform's Nautilus cluster, supported by the National Science Foundation under Award Numbers CNS-1730158, ACI-1540112, ACI1541349, OAC-1826967, the University of California Office of the President, and the University of California San Diego’s California Institute for Telecommunications and Information Technology/Qualcomm Institute.

\bibliography{custom}

\appendix
\section{Prompt Examples}
\label{app:prompt-examples}
\begin{figure*}[h]
    \centering
    \begin{subfigure}[b]{0.49\textwidth}
        \centering
        \includegraphics[width=\textwidth]{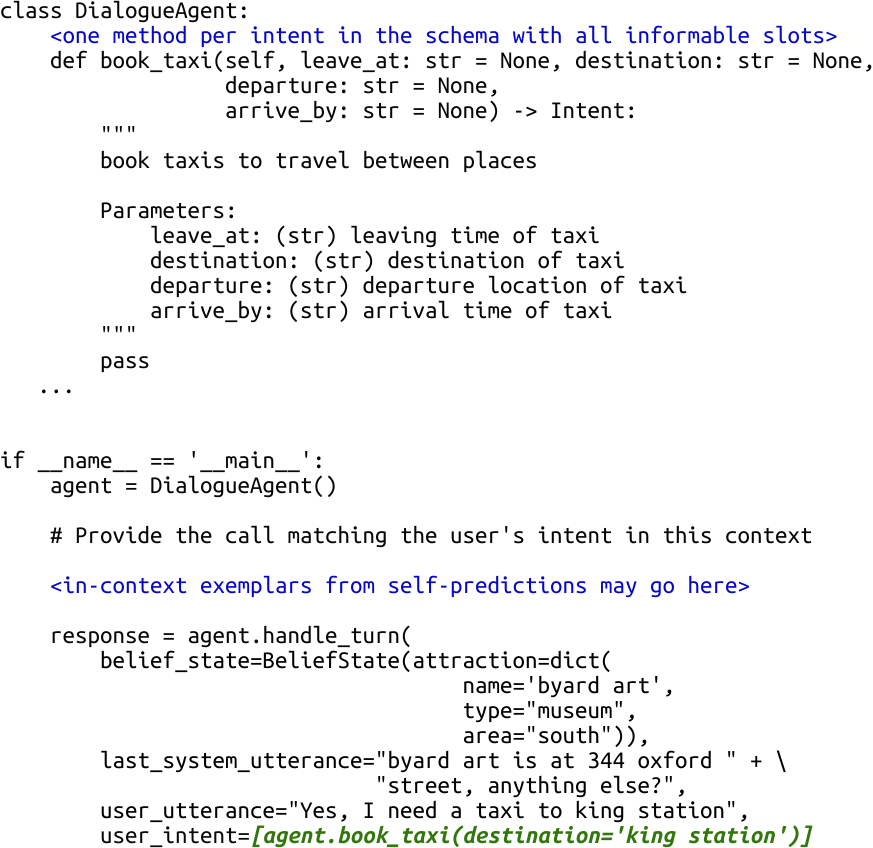} %
        \caption{Our `direct' DST prompt with italicized \dstcolor{\textit{\textbf{completion}}}}
        \label{fig:direct-dst-prompt}
    \end{subfigure}
    \hfill
    \begin{subfigure}[b]{0.49\textwidth}
        \centering
        \includegraphics[width=\textwidth]{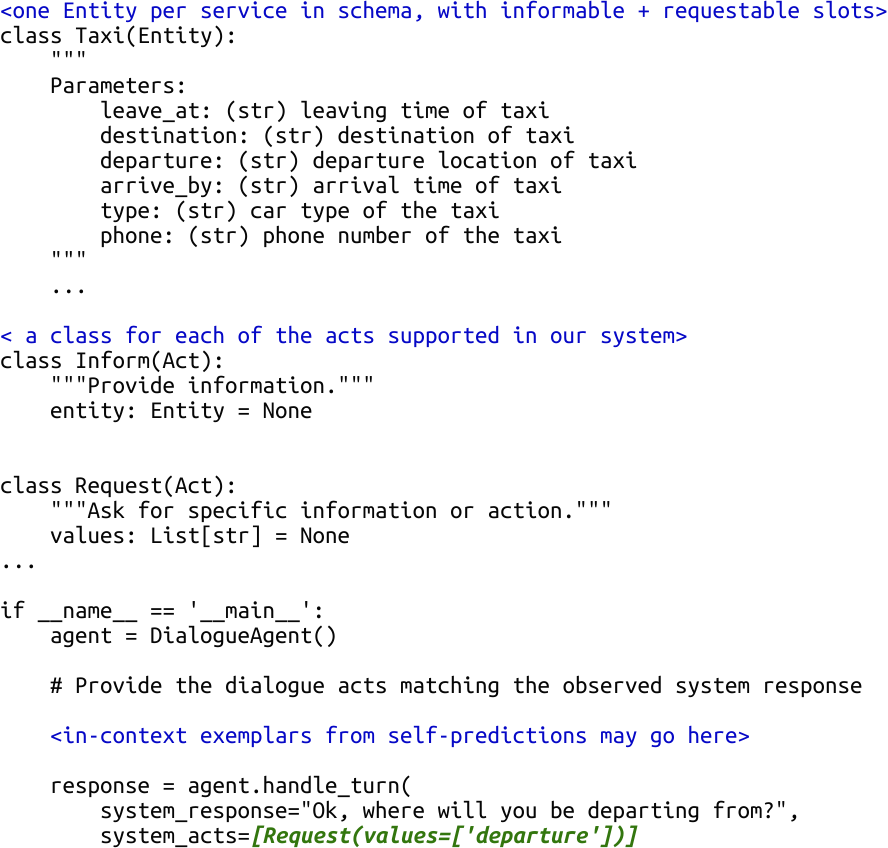} %
        \caption{Our `direct' act tagging prompt, with italicized \dstcolor{\textit{\textbf{completion}}}}
        \label{fig:direct-act-tag-prompt}
    \end{subfigure}
    \caption{Abridged prompt and completion examples from our in-context learning approach to initial labelling for DST and DAT (Act Tagging), best viewed in color. Key-word arguments are used to include variables from the turn context and to prefix the completion}
    \label{fig:direct-prompt-examples}
\end{figure*}

\autoref{fig:direct-prompt-examples} provides abridged instances of our direct prompts for DST and for Act Tagging. \autoref{fig:direct-dst-prompt} shows our prompt for inferring API call(s) or changes to the dialogue state from an unlabeled dialogue, as detailed in \autoref{sec:methods-dst}. Our prompts use python keyword arguments to provide the input variables for a given sub-task, and to prompt the LLM for the next variable of interest. Using the arbitrary ordering of keyword arguments in Python function calls, our `channel' prompts simply re-order the arguments in order to score the likelihood of the user's utterance given the predicted state change.
\autoref{fig:direct-act-tag-prompt} provides a similar abridged instance of our direct prompt for tagging dialogue acts in an unlabeled dialogue. Here, we simply condition on the observed system response $r_t$.

\section{Metric Details}
\label{app:metric_details}
\paragraph{End-to-End (E2E) Dialogue Metrics}
We measure end-to-end dialogue performance using the Inform rate, Success rate, and BLEU, following prior works, using the automatic evaluation provided by \citet{nekvinda-dusek-2021-shades}.\footnote{\href{https://github.com/Tomiinek/MultiWOZ_Evaluation}{https://github.com/Tomiinek/MultiWOZ\_Evaluation}}

A dialogue is considered Informed if the most recently mentioned result for each domain meets the user's goal constraints, and is considered Successful if it is Informed and all values for requested slots are presented to the user. 
For example, if a user were to ask `Can you give me the phone number of a cheap hotel in the east part of town?', the dialogue would be Informed if we refer them to a hotel that is actually in the cheap price range and in the east, and Successful if we additionally provide the phone number, as requested. 
BLEU is computed against a single reference response, and the Combined score is $0.5(\text{Inform} + \text{Success}) + BLEU$.

\paragraph{Dialogue State Tracking Metrics} Following prior works, we evaluate DST performance with joint-goal accuracy (JGA): for a turn $x_t$, a dialogue state prediction $\hat{y}_t$ is considered correct only if all slot names and values match the gold annotation state $y_t$. We again use the evaluation provided in \citet{nekvinda-dusek-2021-shades}. Following their work, we accept fuzzy matches for non-categorical string values, such as the name of a restaurant or hotel, using the \texttt{fuzzywuzzy} library and a fuzz ratio of 0.95.\footnote{\href{https://pypi.org/project/fuzzywuzzy/}{https://pypi.org/project/fuzzywuzzy/}} 

\section{Dialogue Acts}
\label{app:dialogue_acts}

Following \citet{paul19b_interspeech}, we use a universal set of dialogue acts for managing our agents communicative intents. We omit some acts for simplicity and to reduce the context length required to enumerate them in a prompt. \autoref{tab:dialogue_acts} lists each act and a description. Since our dialogue set is not directly comparable to prior works, we do not directly evaluate act tagging or policy accuracy. Instead, acts serve only as an intermediate representation for planning responses in our end-to-end system.

\begin{table*}[]
    \centering
    \begin{tabular}{l p{12cm}}
        \textbf{Act} & \textbf{Description (as used in our prompt)} \\
        \hline
         Inform(x=y) &  Provide information. \\
         Offer(x=y) &  System provides an offer or suggestion based on results. \\
         Confirm(x=y) &  Seek confirmation of something. \\
         Affirm(x=y) &  Express agreement or confirmation. \\
         Negate(x=y) & User or System denies or negates. \\
         NotifySuccess(x=y) & Notify of a successful action or result. \\
         NotifyFailure(x=y) & Notify of an error or failure. \\
         Acknowledge & Acknowledge. \\
         Goodbye & Goodbye. \\
         Greeting & Greeting. \\
         ThankYou & ThankYou. \\
         RequestAlternatives & Ask for other options, alternatives, or any additional user goals. \\
         Request(x=?) & Ask for specific information or action. \\
         
         & 
    \end{tabular}
    \caption{Dialogue acts supported by our system, adapted from the universal dialogue acts proposed in \citet{paul19b_interspeech}. ``x=y" indicates the act can take on arbitrary key-value arguments, and ``x=?" indicates the act takes on one or more unpaired arguments. We reduce the number of acts and lengths of descriptions relative to \citet{paul19b_interspeech} in order to fit within the LMs context length}
    \label{tab:dialogue_acts}
\end{table*}

\section{Offline Labeling Algorithm}
\label{app:offline_algorithm}

\textbf{Algorithm 1} gives our algorithm for pseudo-labeling of unlabeled dialogues.

\begin{algorithm*}
\label{alg:offline_label}
\begin{algorithmic}[1] %
\Procedure{InitialOfflineLabel}{$\D_{train}, \retriever$, $\LM$} %
    \State $\Pool \gets \emptyset$ \Comment{Initialize example pool}
    \State $\mathcal{B} \gets []$ \Comment{Store predictions by dialogue id and turn index}
    \For{$t = 0$ \textbf{to} $\maxunderset{d \in \D_{train}}{|d|}$} \Comment{Loop by increasing turn index}
        \ForAll{$(d_{id}, u_t, r_{t-1}, r_t)$ \textbf{in} $\D_{train}$} \Comment{$d_{id}$ is dialogue ID}
            \State $\hat{b}_{t-1} \gets \mathcal{B}[d_{id}][t-1] \textbf{ or } \emptyset$ \Comment{Fetch $\hat{b}_{t-1}$ if known}
            \State $\hat{b}_t \gets \Call{OfflineDST}{\Pool, \retriever, \hat{b}_{t-1}, r_{t-1}, u_t}$
            \State $\hat{A}_t \gets \Call{OfflineActTag}{\Pool, \retriever, u_t, r_t}$
            \State $\Pool \gets \Pool \cup \{(r_{t-1}, u_t, r_t, \hat{b}_t, \hat{A}_t)\}$ \Comment{Add in-context example for future labeling}
        \EndFor
    \EndFor
\EndProcedure
\Procedure{OfflineDST}{$\Pool, \retriever, \hat{b}_{t-1}, r_{t-1}, u_t$}
    \State $\Ek \gets \retriever(\hat{b}_t \concat r_{t-1} \concat u_t, \Pool)$ \Comment{Retrieve up to $k$ in-context examples}
            \State $\C \gets \Delta b_t \topP P(\prompt(\Ek, \hat{b}_{t-1}, r_{t-1}, u_t))$ \Comment{Sample w/ `direct' prompt}
            \State $\Delta \hat{b}_t \gets \argmax{\Delta b_t \in \C}{P(u_t | \prompt(\Ek, \hat{b}_{t-1}, r_{t-1}, \Delta b_t)}$ \Comment{Re-rank w/ `channel' prompt}
            \State \Return $\hat{b}_{t-1} + \Delta \hat{b}_t$
\EndProcedure
\Procedure{OfflineActTag}{$\Pool, \retriever, u_t, r_t$}
    \State $\Ek \gets \retriever(u_t \concat r_t, \Pool)$ \Comment{Retrieve up to $k$ in-context examples}
            \State $\C \gets A_t  \topP (P(\prompt(\Ek, r_t)))$ \Comment{Sample w/ `direct' prompt}
            \State \Return $ \argmax{A_t\in \C}{P(\Ek, A_t, r_t)}$ \Comment{Re-rank w/ `channel' prompt}
\EndProcedure

\end{algorithmic}
\caption{Our algorithm for initial pseudo-labeling of unlabeled dialogues in $\D_{train}$}
\end{algorithm*}

\section{Further results across EM Steps}
\label{app:additional-ablations}

\begin{figure*}
    \includegraphics[width=\textwidth]{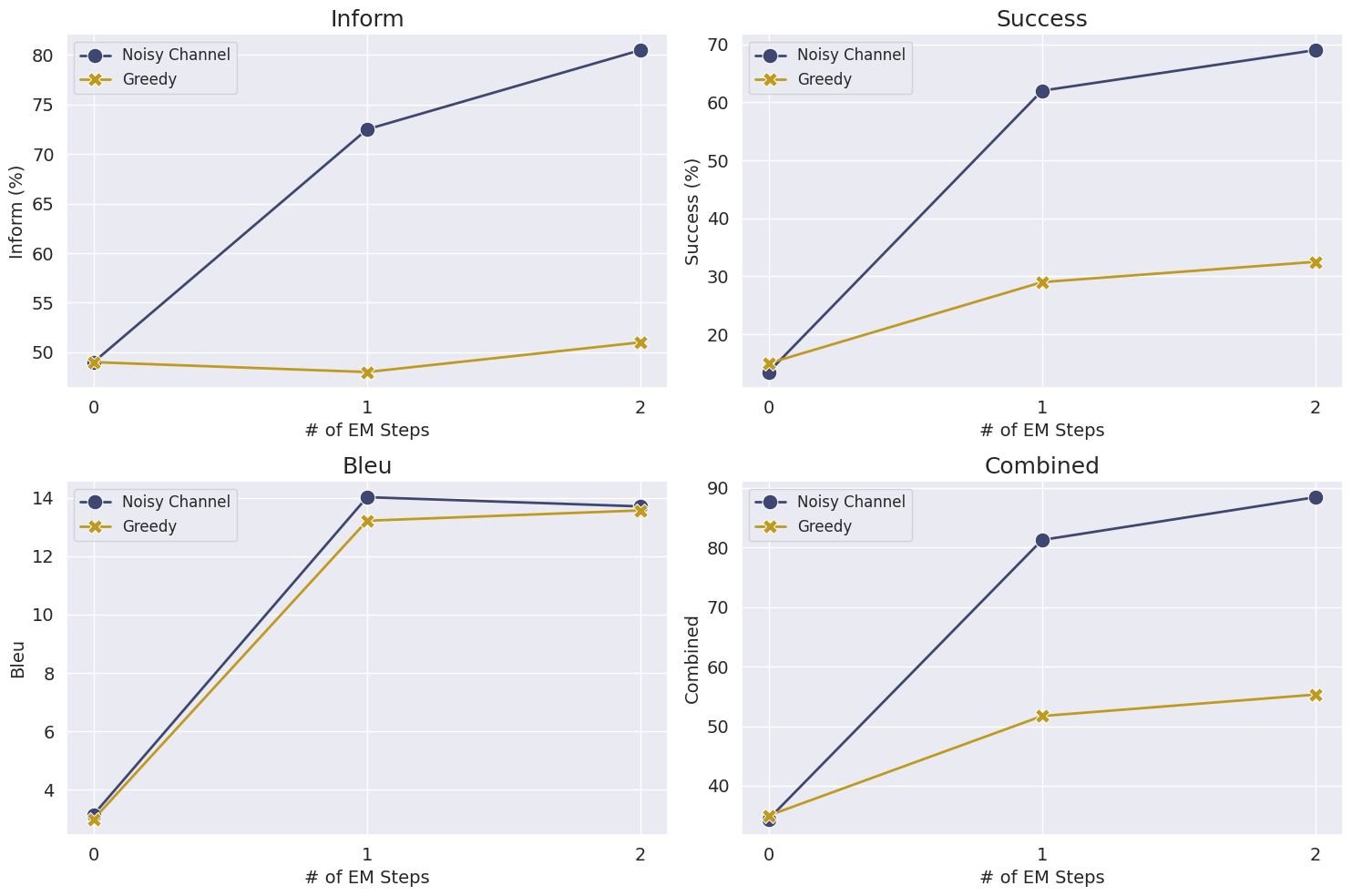}
    \caption{Breaking down $\text{Combined} = 0.5(\text{Inform} + \text{Success}) + BLEU$ into components Inform Rate, Success Rate, and BLEU across iterations of EM between our proposed noisy-channel approach and a greedy ablation, which omits noisy-channel prompting at inference time and when labeling dialogue states \& system acts in the expectation step. We find improvement across all components, and particularly our Inform and Success Rates}
    \label{fig:app-step-plots-grouped}
\end{figure*}

Here we expand on our ablations in \autoref{sec:results}, which evaluates our method with and without our proposed noisy-channel prompting across iterations of expectation-maximization (EM). In \autoref{fig:app-step-plots-grouped}, we break down the performance gains we observed in our `Combined' metric into Inform rate, Success rate, and BLEU, where $\text{Combined} = 0.5(Inform + Success) + BLEU$. `0' iterations of EM indicates our zero-shot prompting system, without any in-context examples or EM. We find that EM substantially improves performance in all cases, and particularly for our noisy-channel prompting approach. We find the noisy channel prompting approach improves performance on all metrics, with the most substantial gains over the greedy baseline in Inform and Success rates. 
This suggests that within our algorithm, noisy-channel inference may be particularly important when inferring the system's dialogue acts in order to reverse-engineer an accurate policy.

\begin{figure*}
    \centering
    \includegraphics[width=0.5\textwidth]{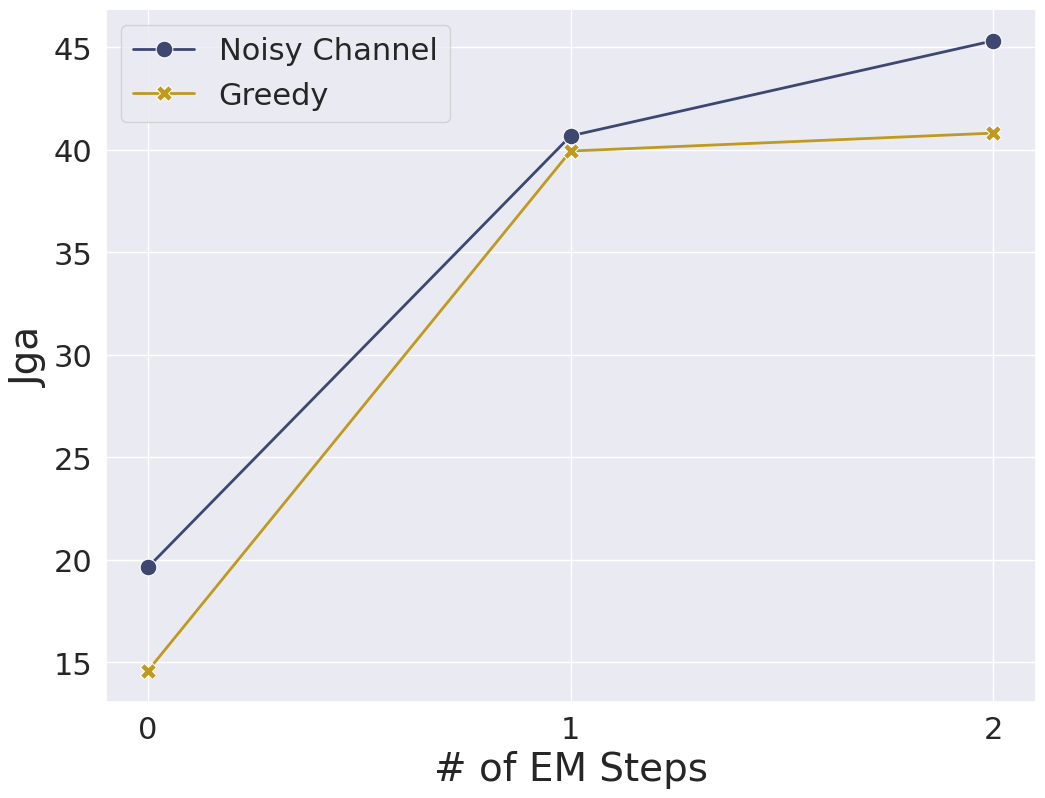}
    \caption{Joint Goal Accuracy (JGA) of our inferred API call(s)/Dialogue states across iterations of EM. We find improved dialogue state tracking performance when using our noisy-channel method at inference time and when labeling dialogue states offline in the expectation step for training, compared to a greedy direct prompting approach}
    \label{fig:app-jga-over-steps}
\end{figure*}

In \autoref{fig:app-jga-over-steps}, we analyze dialogue state tracking performance across iterations of EM using Joint Goal Accuracy (JGA). We find our noisy-channel prompting approach improves the accuracy of our dialogue state tracking predictions across iterations of EM when compared to a greedy, direct prompting approach.

\section{Contamination Search \& Result Details}
\label{app:contamination_details}

\subsection{Procedure} 

We detail our method for finding instances of task contamination within the StarCoder pre-training set.
We are particularly interested in \textit{supervised pairs $(x, y)$} where $y$ belongs to our schema of interest $\schema$, for any of the dialogue sub-tasks used in our system. 
We devise a method for searching the complete pre-training corpus for contaminated $(x, y)$ pairs, where $x$ is an utterance we might observe from either the system or user, and $y$ is the latent dialogue state change or dialogue act supporting $\schema$. 
For each utterance $x$ from either the system or user, we collect all documents from the pre-training corpus which contain the complete utterance. 
We use the elastic search index provided for the StarCoder pre-training data, which accounts for differences in capitalization, punctuation, and interrupting white-space.\footnote{\href{https://github.com/bigcode-project/search/blob/main/index.py}{https://github.com/bigcode-project/search/blob/main/index.py}} 
Following this, we search matching documents for keywords from $y$ (e.g. slot names and values) to determine which of these documents may plausibly contain a supervised label and warrant manual review. 
For dialogue states, these are the slot names and values, discarding extremely generic keywords like `name'.
For act tags, these are the act names, slots, and values.
We then consider a document to need manual review if 40\% or more of the keywords are found in the 500 characters before or after a matching $x$ in a document.
Finally, we hand-check the remaining documents and extract contaminated $(x, y)$ pairs.

\subsection{Examples}
\autoref{tab:contamination-examples} contains examples of contamination discovered in our search process, and the type of document in which they were found. Notably, none of the examples found closely match our output formatting.

\begin{table*}[htbp]
    \centering
    \begin{tabular}{>{\raggedright\arraybackslash}p{0.25\textwidth}|>{\raggedright\arraybackslash}p{0.25\textwidth}|r|r}
        \textbf{Contaminated Input} & \textbf{Contaminated Output} & \textbf{Sub-Task} & \textbf{Source}  \\ 
        \hline
         I need a restaurant to dine at in Cambridge on my upcoming trip . I need info about chiquito restaurant bar restaurant . & restaurant-inform<<<name===chiquito restaurant bar & DST & Jupyter Notebook \\
         \hline 
         i would like to book a 5 star , or closest to it , in the east part of town please . & "<SOB> hotel { area = east, stars = 5, type = hotel } <EOB>
<SOB> hotel { area = east, stars = 5 } restaurant { area = east } <EOB>" & DST & Python  \\
        \hline
                 [Syst] the train id is tr8292 and the price is 16.50 pounds. & [SYS\_DA] train-inform-leave-tr8292
                [SYS\_DA] train-inform-ticket-16.50 pounds & Act Tagging & Github Issue \\
        \hline
    \end{tabular}
    \caption{Example inputs and outputs in contaminated documents from each task, discovered in the StarCoder pre-training corpus. We include the source type of each document}
    \label{tab:contamination-examples}
\end{table*}

\section{Error Analysis}
\label{app:error-analysis}

We conduct an error analysis comparing the greedy model to our best noisy channel model (both with two EM steps). We randomly sampled 10 validation dialogues where at least one model was unsuccessful and analyze each system's behavior. Here we present examples of each error type from \autoref{tab:error-analysis-summary}. Examples can be viewed in \autoref{tab:error-analysis-1}.

The first failure case we present is a dialogue state tracking failure, in which the DST system misses a slot critical to achieving the users goal.

The second example presents a common failure mode in the greedy model, in which commonly occurring dialogue acts are predicted when not appropriate in the context.
In this example, the greedy model predicts that it should `Request' the area of the attraction from the user, even though it is already known. This causes the response generator to hallucinate that there must be no API results to be shared with the user, and that the user should look for attractions in another area (i.e. a `RequestAlternatives').
Our noisy channel model correctly chooses to `Inform' the user, and produces a response faithful to this dialogue act.

The third example presents sub-optimal dialogue acts used by both models. 
The noisy channel and greedy model both predict that they should request information about the attraction `type' from the user, a seemingly reasonable conclusion. Alternatively, the human agent in both cases proactively makes a suggestion. In an online evaluation, the user could provide this information possibly complete a successful dialogue, though in a few more turns than required of the human agent. In the offline evaluation, these dialogues are considered unsuccessful.

In the final example, we show an instance of the greedy model not being faithful to the dialogue acts predicted by the policy. In this case, the predicted dialogue acts are also inappropriate: the system should not request information from the user that they would not be expected to know.
The generated response instead tries to offer a restaurant to the user, but without a value grounded in the API response. This can be considered a hallucination, as the restaurant may or may not exist or match the user's criteria.

\begin{table*}[h]
    \resizebox{\textwidth}{!}{
    \centering
    \begin{tabular}{l}
    \textbf{Error Type: Incorrect Dialogue State} \\
    \hline
    \textbf{User:} find me a place to stay which has 0 star and preferably a guest house \\
    \textbf{System:} There are two guesthouse-type hotels in Cambridge with 0 stars. Do you have an area of town in mind? \\
    \textbf{User:} North part of town for sure \\
    \textbf{System:} that could be city centre north b and b.it is cheap with both internet and parking. can i reserve for you? \\
    \textbf{User:} Yes, please book it for 6 people and 3 nights starting from tuesday. \\
    \textbf{System:} I booked your hotel reservations for 6 people for 3 days starting Tuesday. Your reference number is : EB5RRD33 . \\
    \hspace{3.4em} May i help you with something else. \\
    \textbf{User:} I am looking to go to a multiple sports center too. \\
    \textbf{System:} on what area do you need the attraction? \\
    \textbf{User:} I'd like it to be in the centre, please. \\
    \hdashline
    \textbf{DST Prediction (\redxmark):} \smalltt{\{..., 'attraction': \{'type': 'multiple sports',} \\
    \hspace{18em} \smalltt{\textcolor{red}{- 'area': 'centre'}} \\
    \hspace{19em} \smalltt{\}\}} \\ 
    \\
    \textbf{Error Type: Inappropriate Dialogue Act (Greedy)} \\
    \hline
    \textbf{User:} I'm looking for a train that departs from birmingham new street. It should arrive by 18:30. \\
    \textbf{System:} There is one arrives at 17:23 and another at 18:23. What day are you traveling? \\
    \textbf{User:} I am leaving on tuesday for cambridge. \\
    \textbf{System:} I have train TR5630 leaving at 15:40 and arriving at 18:23. Would that work for you? \\
    \textbf{User:} That sounds great. Thanks. \\
    \textbf{System:} Did you need tickets for that train? \\
    \textbf{User:} Not right now, but thanks for asking. I was wondering if there are any parks on the south side of town, though. \\
    \hdashline
    \textbf{Noisy Channel Acts (\greencmark):} \smalltt{[Inform(entity=Attraction(name=`[value\_name]', area=`[value\_area]'))]} \\
    \textbf{Noisy Channel Response (\greencmark):} There is the \dstcolor{[value\_name]} in the [value\_area] of town. \\
    \textbf{Greedy Acts (\redxmark):} \smalltt{[Request(service=`attraction', values=[`area'])]} \\
    \textbf{Greedy Response (\redxmark):} There are no parks on the south side of town. Would you like me to look in another area? \\
    \textbf{Human Acts}: \smalltt{Offer(entity=Attraction(address=`[value\_address]', area='[value\_area]',} \\\hspace{16.75em} \smalltt{price=`[value\_price]', name=`[value\_name]')} \\
    \textbf{Human Response}: On the [value\_area] we have [value\_name] at [value\_address]. The admission is [value\_price] are you interested? \\ 
    \\
    \textbf{Error Type: Suboptimal Dialogue Act (Both)} \\
    \hline
    \textbf{User:} i need information about wagamama restaurant. \\
    \textbf{System:} i have their info, what would you like to know? \\
    \textbf{User:} I need to book a table for 2 people please. \\
    \textbf{System:} Okay! For when? \\
    \textbf{User:} Wednesday, 17:45. Reference number, please? \\
    \textbf{System:} Let me book that for you now. \\
    \textbf{User:} After you get around to that reference number, I also want to go somewhere in the Centre of town. \\
    \hdashline
    \textbf{Noisy Channel Acts (\redxmark):} \smalltt{[Request(service='attraction', values=['type'])]} \\
    \textbf{Greedy Acts (\redxmark):} \smalltt{[Request(service='attraction', values=['type'])]} \\ 
    \textbf{Noisy Channel Response (\redxmark):} What type of attraction are you looking for? \\ 
    \textbf{Greedy Response (\redxmark):} What type of attraction are you looking for? \\
    \textbf{Human Acts:} \smalltt{[Offer(entity=Attraction(name='all saints church'))]} \\  
    \textbf{Human Response:} What about [value\_name]? \\  
    \\
     \textbf{Error Type: Not Faithful to Act (Greedy)} \\
    \hline
    \textbf{User:} Do you have any recommendations for restaurants in Cambridge? \\
    \textbf{System:} There are 110 restaurants in Cambridge, is there a specific cuisine you're looking for? \\
    \textbf{User:} I'm sorry, it is actually a train that I'm looking for. I need to depart from Cambridge on Saturday. \\
    \textbf{System:} Well can you tell me yur destination please? \\
    \textbf{User:} I need to go to leicester and leave from cambridge after 18:00. There are 8 of us, can you make a reservation \\ \hspace{2.4em} and send me the reference number? \\
    \textbf{System:} I booked you on the TR6034 that leaves Cambridge at 18:21 and arrives at 20:06. Your reference \\ \hspace{3.4em} is NYXZ1SOR and the total fee of 241.92 is payable at the station \\
    \textbf{User:} Perfect. I am also looking for a moderate priced restaurant in the east \\
    \textbf{System:} There are two indian restaurants and one italian. Do either of these interest you? \\
    \textbf{User:} I am interested in Italian. \\
    \hdashline
    \textbf{Noisy Channel Acts (\greencmark):} \smalltt{[Offer(entity=Restaurant(name='[value\_name]'))]} \\
    \textbf{Noisy Channel Response (\greencmark):} I would recommend [value\_name]. Would you like me to book a table for you? \\
        \textbf{Greedy Acts (\redxmark):} \smalltt{[Request(service='restaurant', values=['name', 'address', 'phone', 'postcode'])]} \\
    \textbf{Greedy Response (\redxmark):} I have the caffee uno located in the centre. Would you like me to make reservations? \\
    \textbf{Human Acts:} \smalltt{Inform(entity=Restaurant(address='[value\_address]', name='[value\_name]'))} \\
    \textbf{Human Response:} The [value\_name] is located at [value\_address]. Would you like me to book a reservation? \\
    
    \end{tabular}
    }
    \caption{Examples of each error type from our analysis in \autoref{sec:error-analysis}. In the first error, we show a dialogue state tracking failure which prevents the agent from providing one of the correct attractions. In the second, we show an an inappropriate dialogue act, in which the greedy model requests an area from the user, even though it is already known, leading to a hallucination that there must be no results. In the third, we show a suboptimal dialogue act, used by both systems. Both responses are plausible, but the human agent proactively assists the user in fewer turns. In the final example, we show an instance in which the Greedy model is not faithful to its chosen (incorrect) dialogue act, offering a restaurant that may or may not actually exist or match the user's criteria}
    \label{tab:error-analysis-1}
\end{table*}

\end{document}